\documentclass[10pt,conference,letterpaper]{IEEEtran}

\usepackage[utf8]{inputenc} 
\usepackage[T1]{fontenc}    

\usepackage{url}

\usepackage{cite}
\usepackage{amsmath,amssymb,amsfonts}
\usepackage{nicefrac}       
\usepackage{algorithm}
\usepackage{algpseudocode}
\usepackage[pdftex]{graphicx}
\usepackage{textcomp}
\usepackage{subcaption}

\usepackage[export]{adjustbox}

\usepackage{booktabs} 
\usepackage{multirow}

\usepackage{color}
\usepackage{xcolor}

\usepackage{balance}

\usepackage{xspace}

\usepackage{verbatim}

\usepackage{afterpage}

\usepackage{microtype}      

\def\BibTeX{{\rm B\kern-.05em{\sc i\kern-.025em b}\kern-.08em
    T\kern-.1667em\lower.7ex\hbox{E}\kern-.125emX}}

\newcommand{\myparagraph}[1]{\smallskip \noindent \textbf{#1.}}

\newcommand{\ignore}[1]{}

\begin{document}


\title{Are Self-Driving Cars Secure? Evasion Attacks against Deep Neural Networks for Steering Angle Prediction}


\author{\IEEEauthorblockN{Alesia Chernikova\IEEEauthorrefmark{1},
Alina Oprea\IEEEauthorrefmark{1},
Cristina Nita-Rotaru\IEEEauthorrefmark{1} and
BaekGyu Kim\IEEEauthorrefmark{1}\IEEEauthorrefmark{2}}
\IEEEauthorblockA{\IEEEauthorrefmark{1}Northeastern University, Boston, MA}
\IEEEauthorblockA{\IEEEauthorrefmark{2}Toyota ITC, USA}}


\maketitle


\begin{abstract}

Deep Neural Networks (DNNs) have tremendous potential in advancing the vision for self-driving cars. However, the security of DNN models in this context leads to major safety implications and needs to be better understood. We consider the case study of
steering angle prediction from camera images, using the dataset from the 2014 Udacity challenge. We demonstrate for the first time adversarial testing-time attacks for this application for both classification and regression settings. We show that minor modifications to the camera image (an $L_2$ distance
of 0.82 for one of the considered models) result in mis-classification of an image to any class of
	attacker’s choice. Furthermore, our regression attack results in a significant increase in Mean Square Error (MSE) -- by a factor of 69 in the worst case.\footnote{Preprint of the work accepted for publication at the IEEE Workshop on the Internet of Safe Things, San Francisco, CA, USA, May 23, 2019.}




\end{abstract}

\section{Introduction}


Advances in Machine Learning (ML) and Deep Neural Networks (DNNs) bring tremendous potential to make autonomous vehicles a reality.  In this setting, sensors such as camera, light detection and ranging sensor (LiDAR), and Infrared (IR) generate streams of real-time data. Envisioned ML applications include: predicting road conditions by interacting with other cars; recognizing risky road conditions; and assisting drivers in taking safer decisions.  For this highly-critical application, safety is the major concern, but unfortunately ML algorithms are not traditionally designed and evaluated from this perspective.


At the same time, the security of ML models at both training and testing time has received lately a lot of attention. Initially, adversarial attacks against supervised learning have been mostly studied in the context of image classification systems~\cite{Biggio13,Szegedy14,Goodfellow14}. But recently these attacks have been extended to other domains, including cyber security~\cite{Srndic14} and speech recognition~\cite{AudioAdv}. To the best of our knowledge, though, adversarial attacks for self-driving cars have not been addressed so far.


In this paper, we demonstrate that classification and regression models for self-driving car applications are also vulnerable to adversarial evasion attacks at testing time.  We consider the case study of steering angle predicting from camera images, using the dataset from the 2014 Udacity challenge 2~\cite{UdacityChallenge}. First, we adapt the state-of-the-art Carlini and Wagner 2017 evasion attack~\cite{Carlini17} to the classification problem of predicting steering direction, using architectures inspired by two Convolutional Neural Network (CNN) models that obtained good results in the Udacity challenge~\cite{UdacitySDC,bojarski2016end}. We show that minor modifications to the camera image (an $L_2$ distance of 0.82 for one of the considered models) result in mis-classification of an image to any class of attacker's choice. Second, we design the first testing-time attack for regression based on CNNs and test them in the setting of this application. We show that our attacks cause significant degradation to the Mean Square Error (MSE) metric used to evaluate regression. In particular, our attack increases the MSE of 10\% of the images by a factor of more than 20 compared to the setting without attack.

Our work calls for further research into the safety implications of these attacks in the self-driving car application domain. As connected cars become more autonomous and new technologies are developed for assisting drivers on the road, it becomes of paramount importance to understand in depth the security and safety of deep learning in this setting.




\section{Background and Threat Model}

\myparagraph{Background on connected cars} Modern cars are outfitted with Electronic Control Units (ECUs) to control specific functions on the car, such as the engine~\cite{woo2015practical} and controlling brakes~\cite{kleberger2011security}. In connected cars, some of these ECUs communicate outside of the car, such as for the infotainment system, remote firmware patching, or on-board diagnostics~\cite{kleberger2011security}. While this adds functionality, it also opens them up to attack. Furthermore, autonomous vehicles replace human-made decisions with decisions made using sensor input from extra cameras, LiDAR, RADAR, etc. These sensors communicate with the control systems via their ECUs, over the CAN bus. In a setting where ECUs have been compromised, lack of authentication on the CAN bus makes it possible for a compromised ECU to send messages as other sensors, such as the camera.

\myparagraph{Neural networks} A feed-forward neural network is a function $y = F(x)$ from input data points $x \in R^n$ to output $y \in R^m$ that depends implicitly on model parameter $\theta$. A neural network has $L$ layers, and the output $F$ is computed by applying a function at each layer. Each layer has a number of output neurons. In each layer, a linear matrix multiplication is followed by a non-linear activation function. For multi-class classification, the last layer uses a softmax activation function with the number of neurons equal to the number of classes.  The inputs to the softmax function are called $\mathsf{logits}$. We define $F$ to be the full neural network including the $\mathsf{softmax}$ function, and
$Z(x) = z$ to be the output of all layers except the $\mathsf{softmax}$, thus $y = F(x) = \mathsf{softmax}(Z(x))$.

Convolutional neural networks (CNNs) are a particular type of feed-forward networks, with the requirement that at least one of the layers performs a \emph{convolution operation} followed by a non-linear activation. A convolution is a linear operation that slides a filter of small size over the output of the previous layer and computes repeatedly dot products of the filter with regions of the input data.



\myparagraph{Udacity challenge} In the Udacity challenge 2~\cite{UdacityChallenge}, the goal is to predict the appropriate steering angle using only imagery from the car's center camera. A negative steering angle implies turning left, while a positive one results in a right turn.  The full dataset consists of 33,608 images and their corresponding steering angle values, in total 70GB of data.



\myparagraph{Threat model} We consider an attacker who is capable of controlling one or multiple ECUs. From here, lack of authentication on the CAN bus can allow an adversary to spoof messages from the camera~\cite{kleberger2011security}. The attacker can modify the image sent by a camera, constructing an adversarial example which will be misclassified by a steering angle controller for the autonomous vehicle. We are concerned with an active attacker with partial control of one or several car ECUs, interested in generating a stealthy perturbation to images produced by the camera. The reasons for which the attack wishes to remain stealthy are multi-fold: (1) to avoid suspicion by humans looking at the camera; (2) to avoid detection by anomaly detection software for threat detection~\cite{jagielski2018threat}. We consider the strongest threat model (\emph{white-box attacks}), which provides the attacker full knowledge of the ML system.



\begin{table}[!thb]
\begin{minipage}{\linewidth}
\begin{tabular}{ll}
\cmidrule(r){1-2}
Layer   & Architecture and Hyper-parameters  \\
\midrule
Convolutional + ReLU & 32 filters of size $3 \times 3  \times 3$\\
MaxPooling & Filter $2 \times 2$     \\
Dropout    & Fraction 0.25       \\
Convolutional + ReLU & 64 filters of size  $3 \times 3 \times  32$\\
MaxPooling & Filter $2 \times 2$     \\
Dropout    & Fraction 0.25       \\
Convolutional + ReLU & 128 filters of size $3 \times 3 \times 64$\\
MaxPooling & Filter $2 \times 2$     \\
Dropout    & Fraction 0.5       \\
Fully-Connected + ReLU & Neurons 1024\\
Dropout    & Fraction 0.5       \\
Fully-Connected + Softmax & Neurons 3\\
\bottomrule
\end{tabular}
\caption{Epoch Model Architecture}
\label{tab:epoch}

\end{minipage}\\

\begin{minipage}{\linewidth}

\begin{tabular}{ll}
\cmidrule(r){1-2}
Layer   & Architecture and Hyper-parameters  \\
\midrule
Batch Normalization Layer & \\
Convolutional + ReLU & 24 filters of size $5 \times 5 \times 3$\\
Convolutional + ReLU & 36 filters of size $5 \times 5 \times 24$ \\
Convolutional + ReLU & 48 filters of size $5 \times 5 \times 36$\\
Convolutional + ReLU & 64 filters of size $3 \times 3 \times 48$ \\
Convolutional + ReLU & 64 filters of size $3 \times 3 \times 64$ \\
Fully-Connected + ReLU & Neurons 582\\
Fully-Connected + ReLU & Neurons 100\\
Fully-Connected + ReLU & Neurons 50\\
Fully-Connected + ReLU & Neurons 10\\
Fully-Connected + Softmax & Neurons 3\\
\bottomrule
\end{tabular}
\caption{NVIDIA Model Architecture}
\label{tab:nvidia}
\end{minipage}
\end{table}

\begin{figure}[h]
\begin{minipage}{\linewidth}
\centering
\includegraphics[width=0.7\columnwidth]{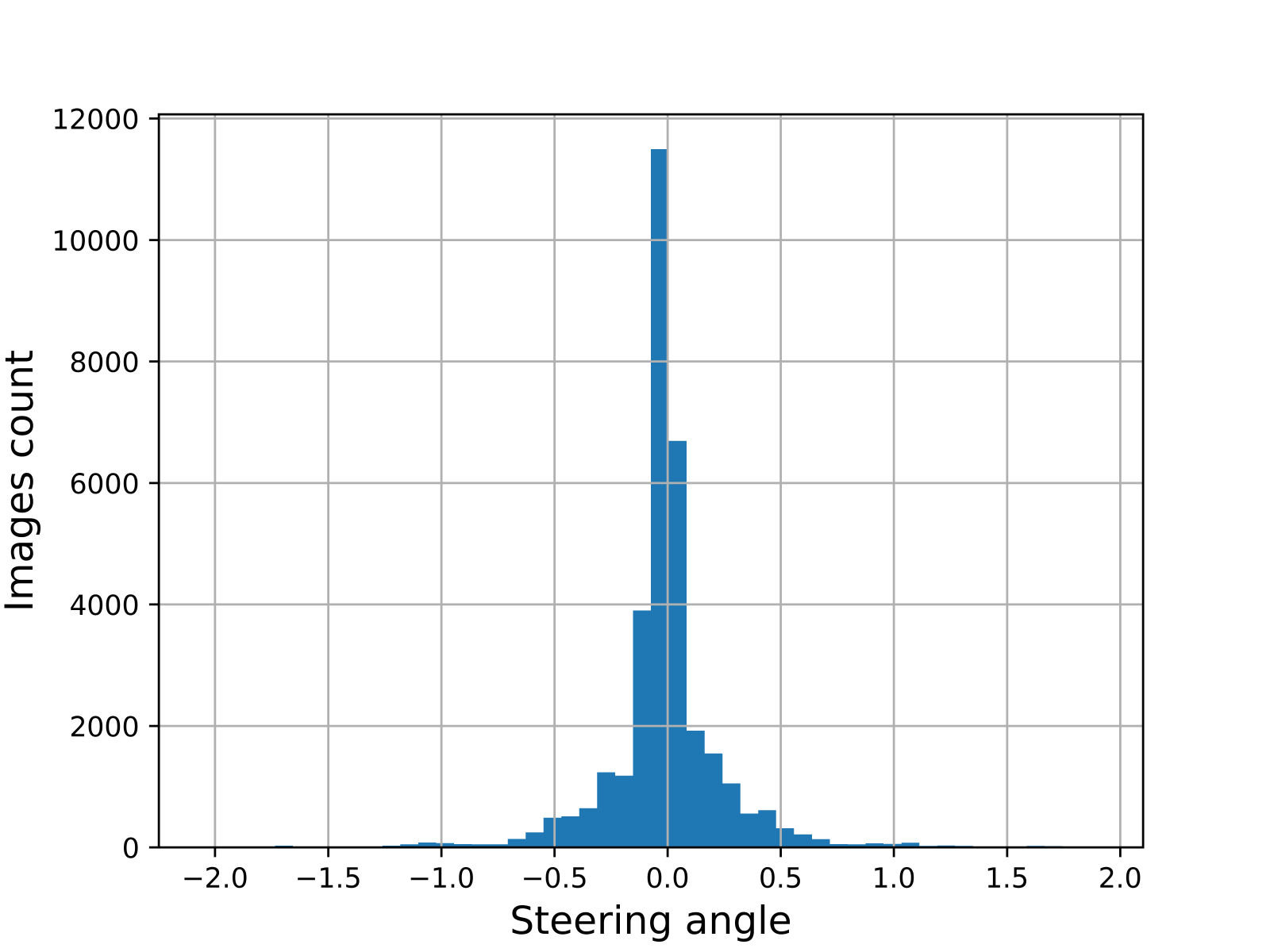}
\caption{Scaled steering angle histogram}
\label{fig:hist}
\end{minipage}
\begin{minipage}{0.45\linewidth}
\captionsetup{type=table}
\begin{tabular}{ll}
\cmidrule(r){1-2}
Statistic    &  Value\\
\midrule
Minimum   &  -2.05\\
Maximum  & 1.9 \\
Mean    &  -0.008\\
Std. dev.     &   0.27\\
\bottomrule
\end{tabular}
\caption{Scaled steering angle distribution}
\label{tab:angle}
\end{minipage}
\begin{minipage}{0.45\linewidth}
\centering
\captionsetup{type=table}
\begin{tabular}{ll}
 \cmidrule(r){1-2}
Parameter    & Value   \\
\midrule
Learning rate & 0.01  \\
Momentum    & 0.9  \\
Batch size    & 128  \\
Epochs   & 50     \\
\bottomrule
\end{tabular}
\caption{Training hyper-parameters}
\label{tab:params}
\end{minipage}
\end{figure}

\section{Attack Algorithm}
In this section we describe the evasion attack against DNNs for steering angle prediction.

\begin{figure*}[t]
\centering
\begin{subfigure}[t]{0.7\columnwidth}
\includegraphics[width=\columnwidth]{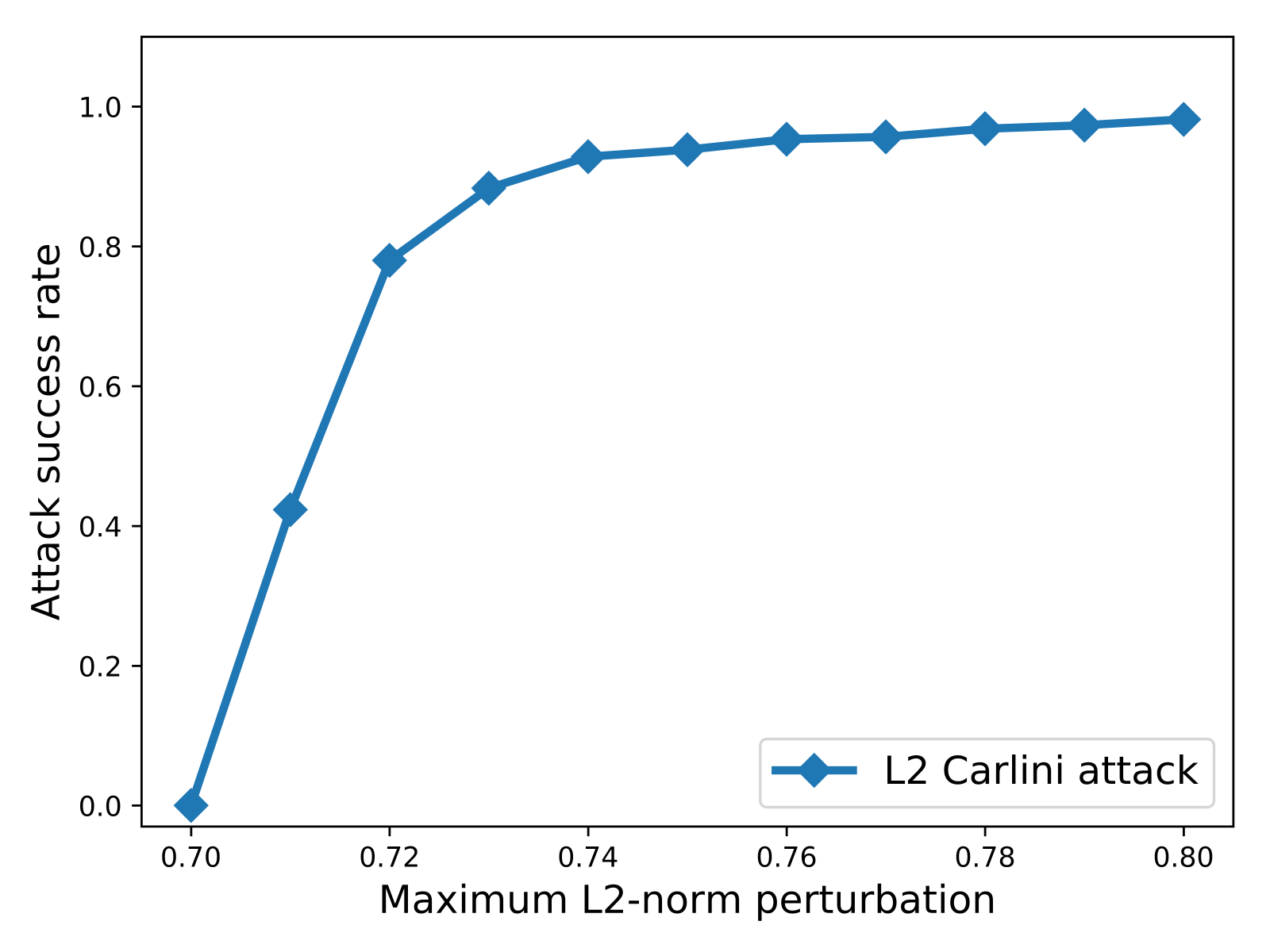}
\caption{Epoch model}
\label{fig:dist1}
\end{subfigure}
\begin{subfigure}[t]{0.7\columnwidth}
\includegraphics[width=\columnwidth]{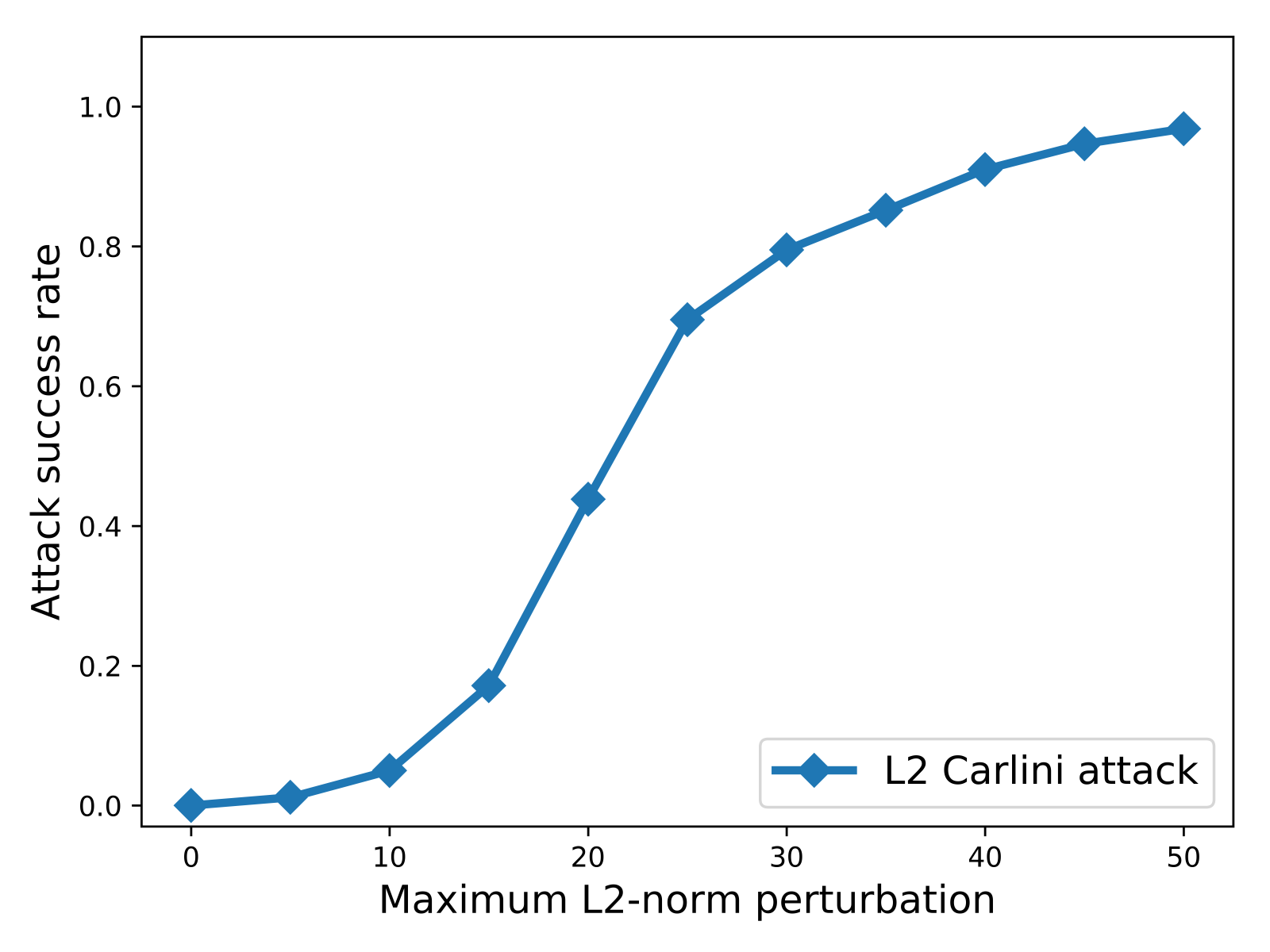}
\caption{NVIDIA model}
\label{fig:dist2}
\end{subfigure}
\caption{Success of attack with respect to distance}
\end{figure*}

\begin{figure*}[h]
\centering
\begin{subfigure}[t]{0.7\columnwidth}
\includegraphics[width=\columnwidth]{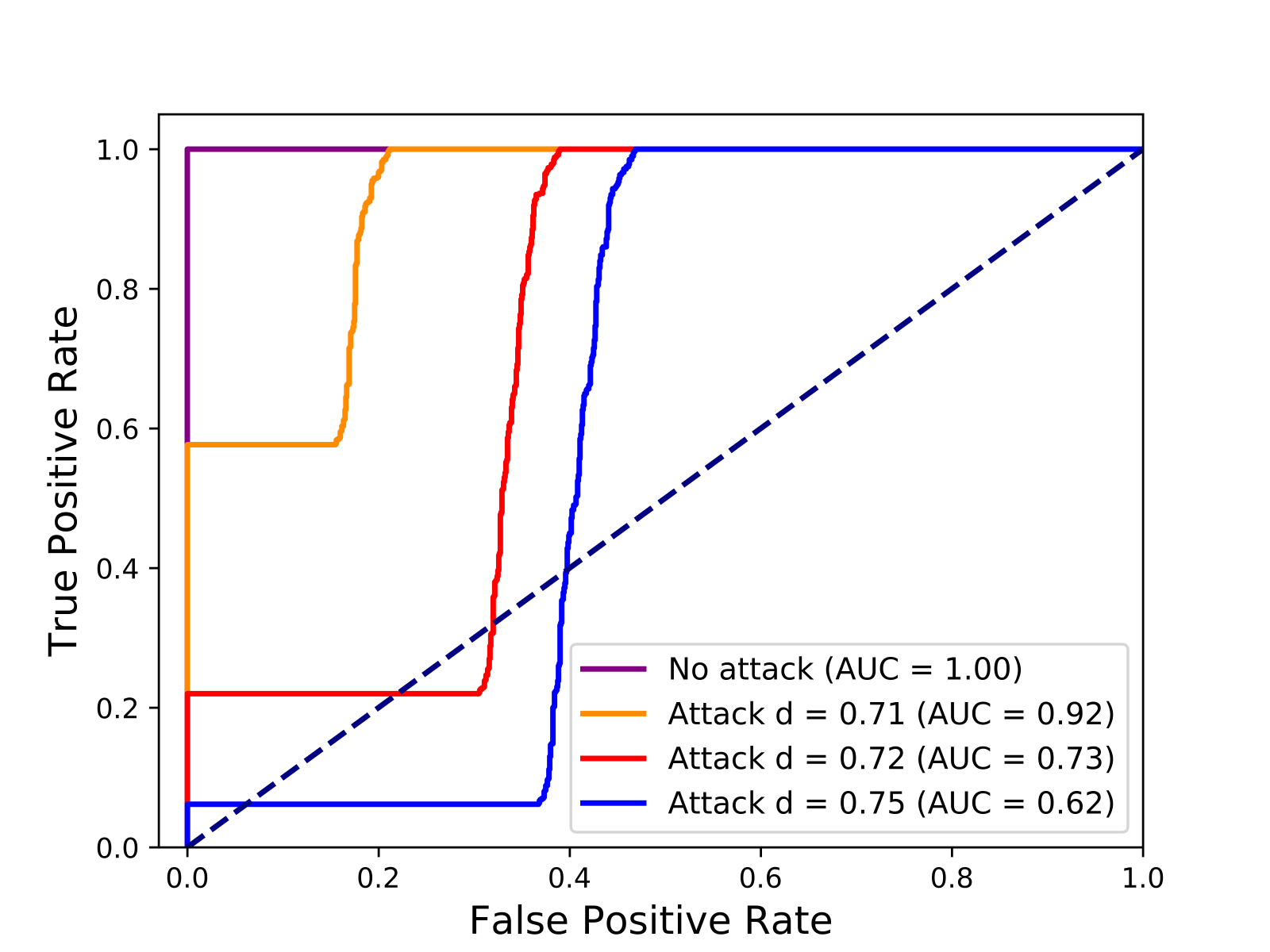}
\caption{Epoch model}
\label{fig:ROC1}
\end{subfigure}
\begin{subfigure}[t]{0.7\columnwidth}
\includegraphics[width=\columnwidth]{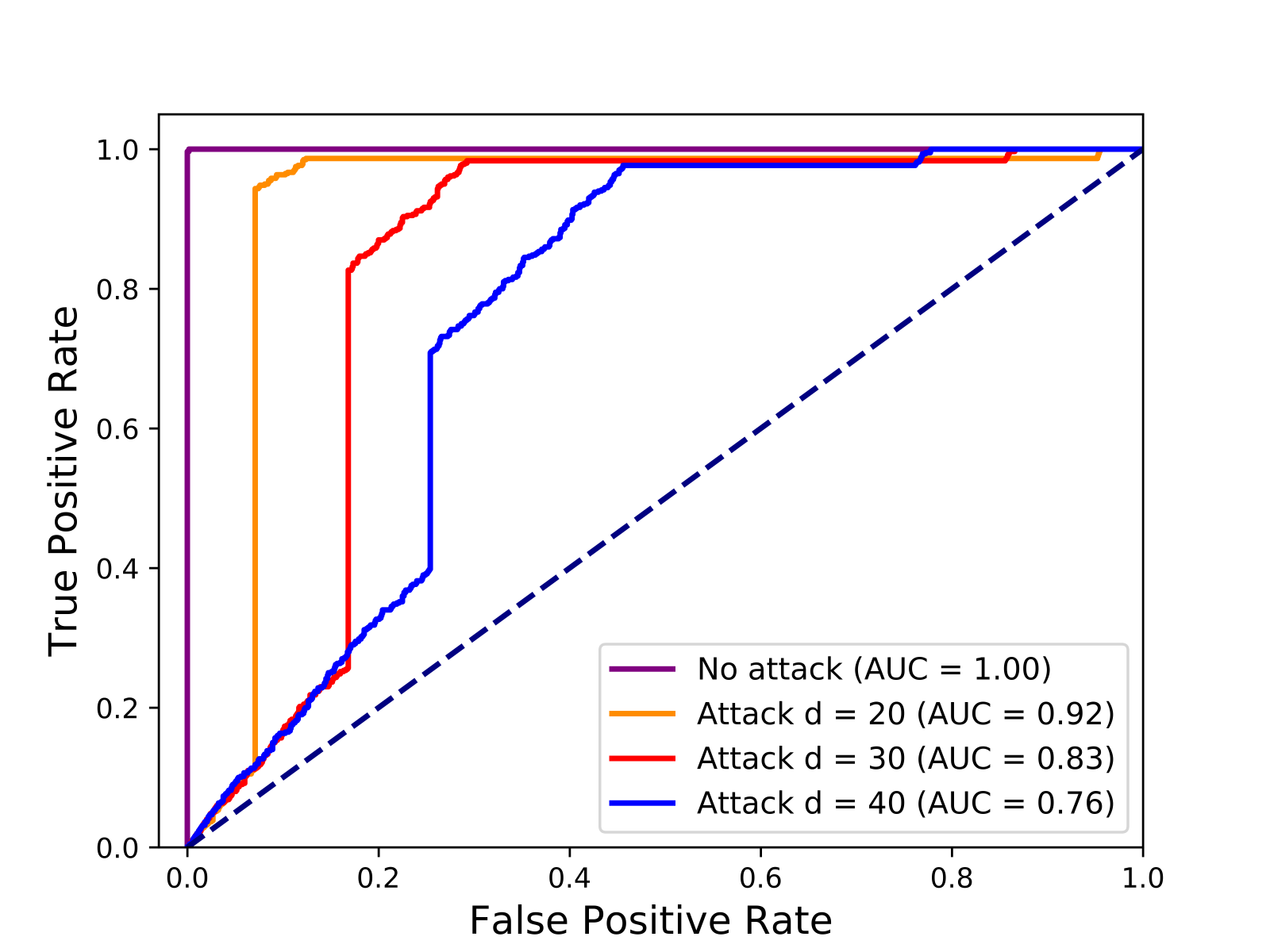}
\caption{NVIDIA model}
\label{fig:ROC2}
\end{subfigure}
\caption{ROC curves for models with and without the attack.}
\end{figure*}

\myparagraph{DNN architectures}
A number of DNN models submitted to the Udacity challenge 2 successfully predict steering angle values, therefore solving the regression problem.

We first consider the classification problem of \emph{predicting the car direction} needed for predicting lane changes and eventually the full vehicle trajectory~\cite{Thrun06}. Based on domain expert recommendation, we select an angle threshold and replace the exact value of the predicted angle with a class: {\bf right} if the steering angle exceeds the positive value of the threshold; {\bf left} if the steering angle is below the negative value of the threshold; and {\bf straight} otherwise. The car direction prediction task takes as input the image camera and predicts the direction the car should take. Second, we consider the regression problem of predicting steering angles, identical to the original challenge problem.




We select two Convolutional Neural Network models for both the classification and regression problems. The first is the \emph{Epoch model}~\cite{UdacitySDC} (also used by DeepTest~\cite{tian2018deeptest}), while the second is inspired by Bojarski et al~\cite{bojarski2016end} (called \emph{NVIDIA model}). The Epoch model consists of 3 convolutional layers, and 2 fully-connected layers (see Table~\ref{tab:epoch}). The NVIDIA model  has the same number of convolutional layers and fully-connected layers as Bojarski et al.~\cite{bojarski2016end}, but less hidden units in the first fully-connected layer to speed up training (see Table~\ref{tab:nvidia}). We adapted both models for classification by adding a last layer with 3 hidden units and softmax activation function. The architecture for regression is similar, excluding the last softmax layer. The NVIDIA model is more complex (467 million parameters) compared to the Epoch model (25 million parameters).

\myparagraph{Evasion attacks against direction classification} We use the $L_2$ distance between the original and adversarial image to measure the amount of perturbation introduced by the attack. In this setting,  the attacker adds negligible perturbations to all image pixels. We also assume that image pixels are normalized in $[0,1]$. We leverage and adapt the state-of-art $L_2$ attack by Carlini and Wagner~\cite{Carlini17}, proposed originally in the context of image classification. The attack crafts adversarial examples by solving the following optimization problem for an image $x$ with original class $i$ to find the perturbation $\sigma$ that transforms it into a targeted class $t \neq i$:
\begin{center}
minimize $||\sigma||_2 + c \times f(x + \sigma)$\\
such that $x + \sigma \in [0, 1]^ d$ \\

$f(x+ \sigma) = (max (Z(x+\sigma)_{j \neq t}) -Z(x+\sigma)_t)^+ $\\
$i$ - original class, $t \neq i$ - adversarial target class.

\end{center}

Here $s^+$ is the notation for $\max(s,0)$, The main intuition is that the optimization objective includes two terms: a distance norm of the adversarial perturbation and a loss function that is minimized when the modified image is classified to the target class $t \neq i$. The hyper-parameter $c$ controls the tradeoffs between the amount of perturbation to the image and the attack success of classifying to the target class.



\begin{figure*}
\centering
\begin{subfigure}[t]{0.6\columnwidth}
\includegraphics[width=\columnwidth]{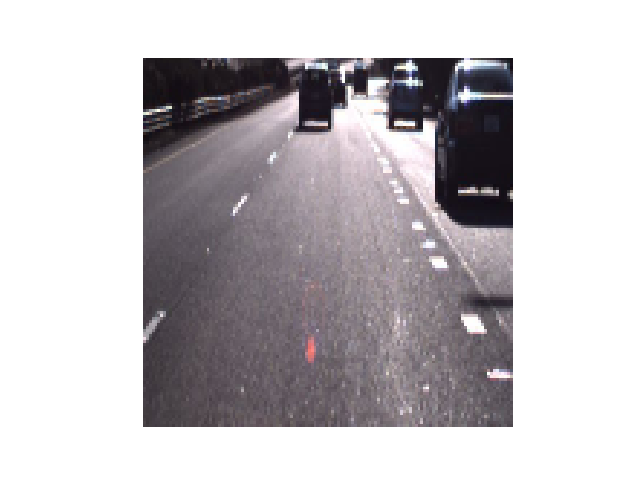}
\caption{Input image,  'straight'}
\end{subfigure}
\begin{subfigure}[t]{0.6\columnwidth}
\includegraphics[width=\columnwidth]{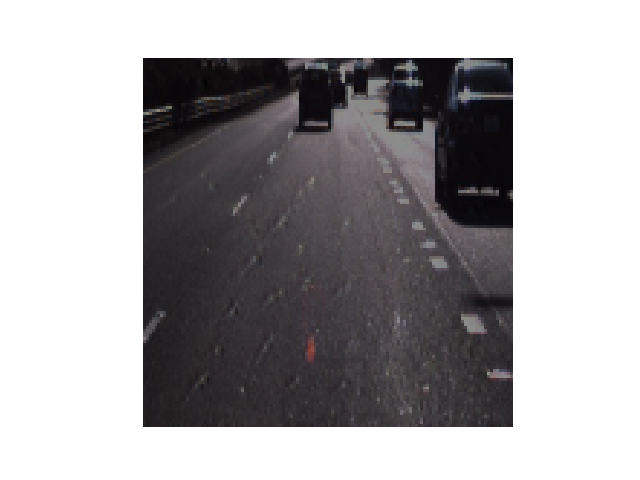}
\caption{Adversarial image, 'left'}
\end{subfigure}
\begin{subfigure}[t]{0.6\columnwidth}
\includegraphics[width=\columnwidth]{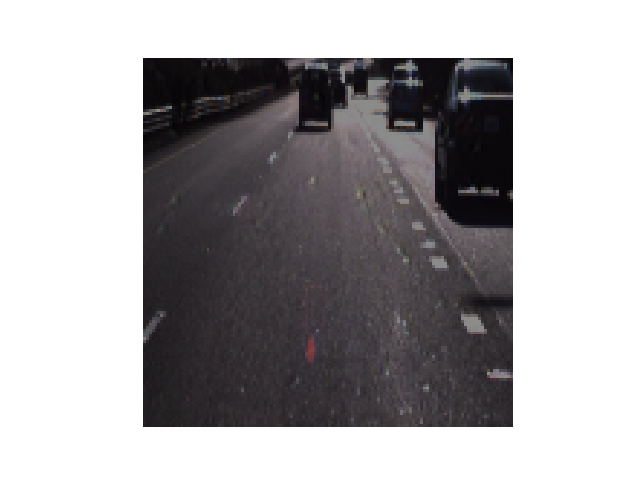}
\caption{Adversarial image, 'right'}
\end{subfigure}

\begin{subfigure}[t]{0.6\columnwidth}
\includegraphics[width=\columnwidth]{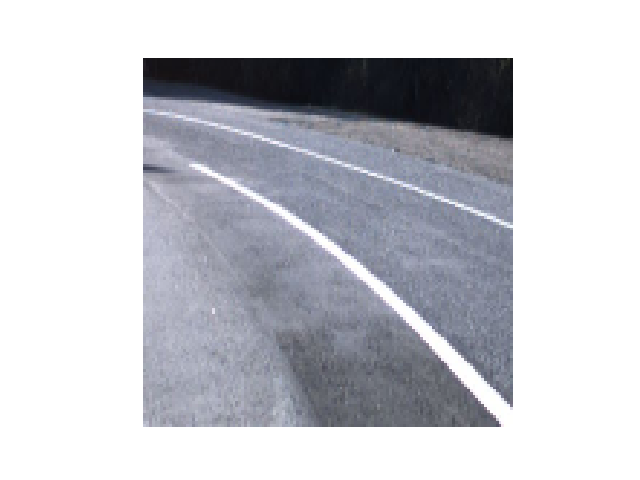}
\caption{Input image,  'left'}
\end{subfigure}
\begin{subfigure}[t]{0.6\columnwidth}
\includegraphics[width=\columnwidth]{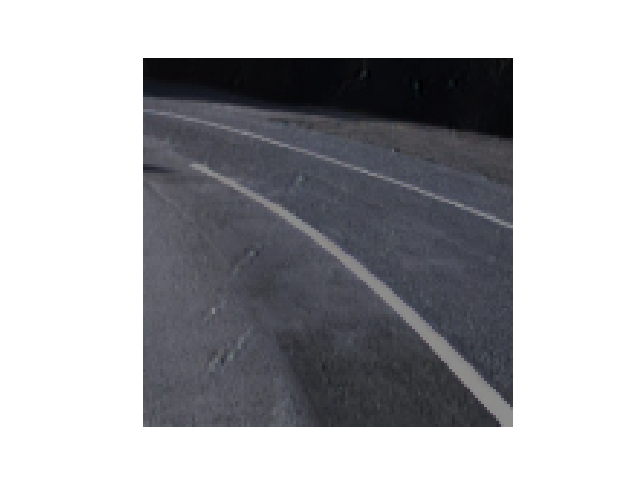}
\caption{Adversarial image, 'straight'}
\end{subfigure}
\begin{subfigure}[t]{0.6\columnwidth}
\includegraphics[width=\columnwidth]{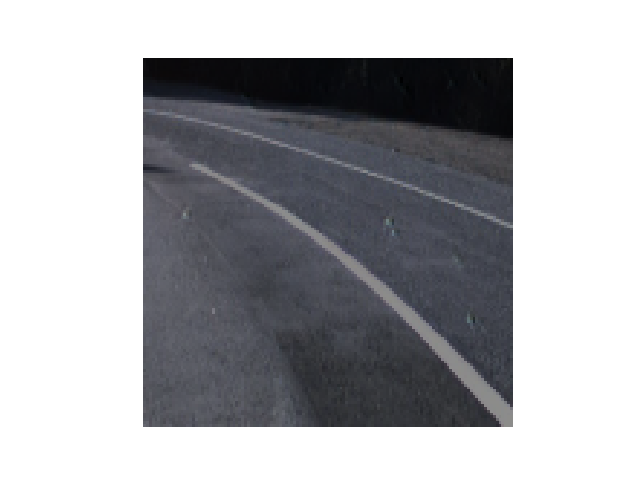}
\caption{Adversarial image, 'right'}
\end{subfigure}

\begin{subfigure}[t]{0.6\columnwidth}
\includegraphics[width=\columnwidth]{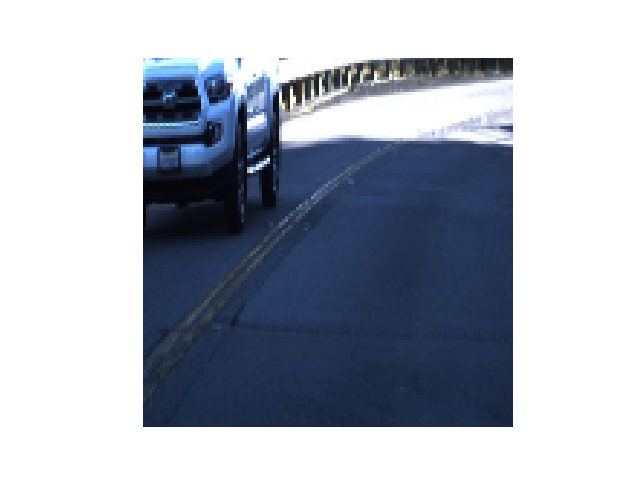}
\caption{Input image,  'right'}
\end{subfigure}
\begin{subfigure}[t]{0.6\columnwidth}
\includegraphics[width=\columnwidth]{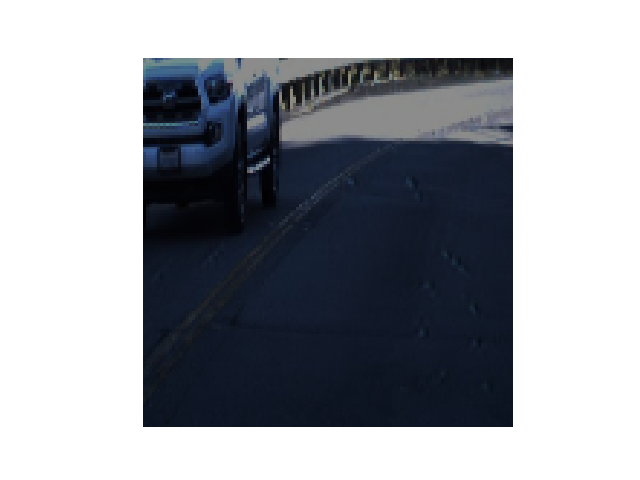}
\caption{Adversarial image, 'straight'}
\end{subfigure}
\begin{subfigure}[t]{0.6\columnwidth}
\includegraphics[width=\columnwidth]{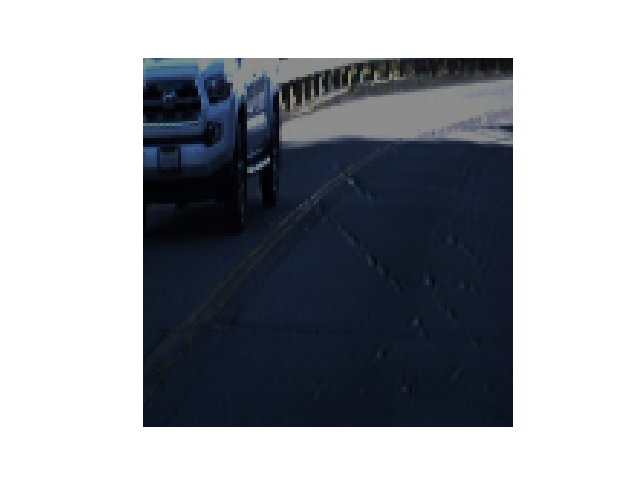}
\caption{Adversarial image, 'left'}
\end{subfigure}
\caption{Adversarial images for the Epoch classification model.}
\label{fig:exepoch}
\end{figure*}

\myparagraph{Evasion attacks against steering angle regression prediction} We are not aware of existing evasion attacks against CNNs for regression. A regression model is typically evaluated by the Mean Square Error (MSE) metric, defined either for single points or over an entire dataset. MSE of a single point $x$ with response $y \in R$ measures the squared residual (e.g., difference between the true response $y$ and the predicted response $\hat{y}=F(x)$). For a dataset, MSE is the average of the squared residuals of all points.  Our main insight is to adapt the classification attack by changing the objective function \emph{to maximize the MSE difference between the predicted response on the adversarial image $F(x+\sigma)$ and the true response $y$}. This way, the attacker attempts to change the prediction on the adversarial image further away from the true value.


Thus, in order to find the adversarial image for original image $x$ with response $y$, the attacker solves the following optimization task with respect to the parameter $\sigma$:

\begin{center}
minimize $||\sigma||_2 - c \times g(x + \sigma, y)$\\
such that $x + \sigma \in [0, 1]^ d$ \\
$g(x+ \sigma, y) = (F(x + \sigma) - y)^2 $
\end{center}

Here $c$ is a hyper-parameter that is found by binary search; it controls the tradeoff between minimizing the image perturbation versus maximizing the MSE value. Small values of parameter $c$ should be used when the goal of making the resulting perturbation negligible is more important than increasing the MSE of the adversarial image.



\section{Experiments}
\label{sec:exp}






\begin{figure*}
\centering
\begin{subfigure}[t]{0.6\columnwidth}
\includegraphics[width=\columnwidth]{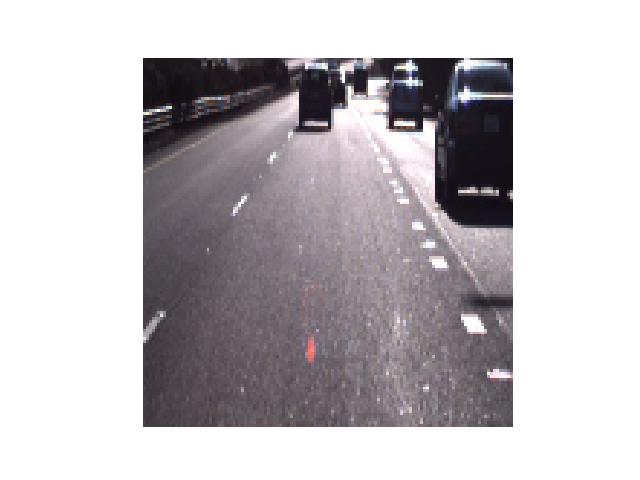}
\caption{Input image,  'straight'}
\end{subfigure}
\begin{subfigure}[t]{0.6\columnwidth}
\includegraphics[width=\columnwidth]{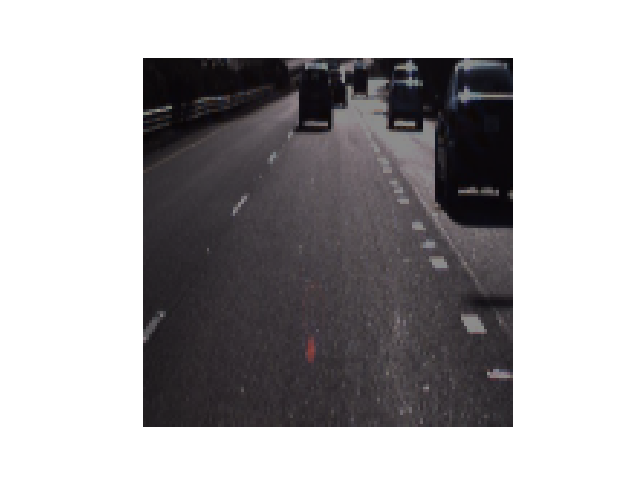}
\caption{Adversarial image, 'left'}
\end{subfigure}
\begin{subfigure}[t]{0.6\columnwidth}
\includegraphics[width=\columnwidth]{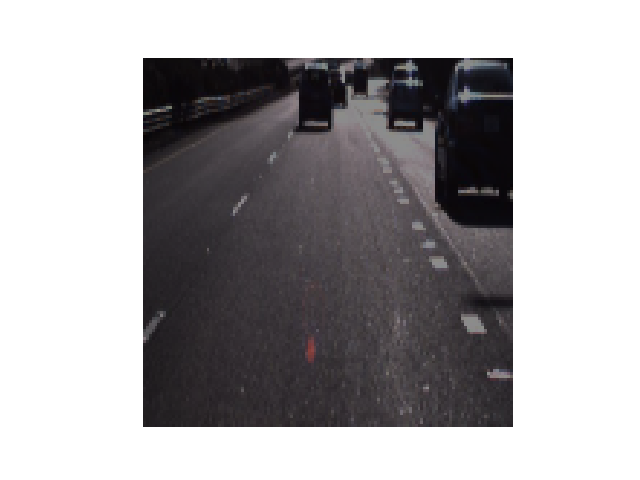}
\caption{Adversarial image, 'right'}
\end{subfigure}

\begin{subfigure}[t]{0.6\columnwidth}
\includegraphics[width=\columnwidth]{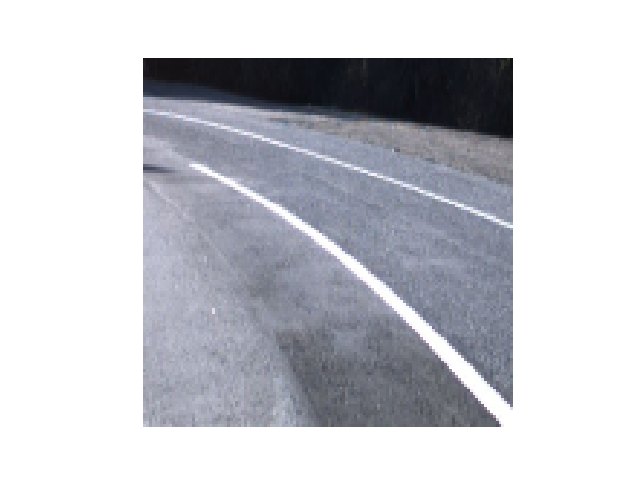}
\caption{Input image,  'left'}
\end{subfigure}
\begin{subfigure}[t]{0.6\columnwidth}
\includegraphics[width=\columnwidth]{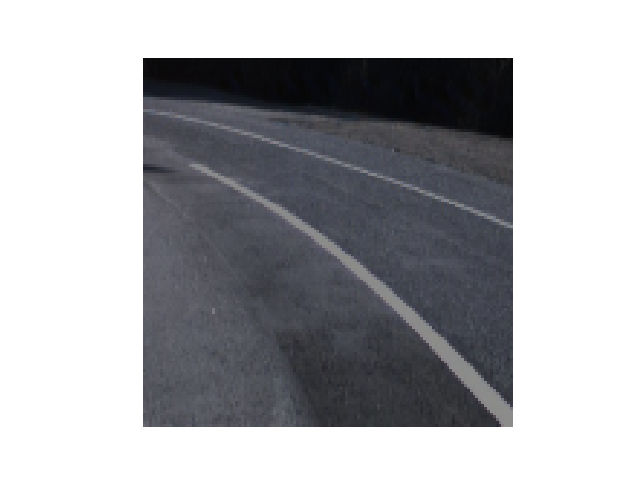}
\caption{Adversarial image, 'straight'}
\end{subfigure}
\begin{subfigure}[t]{0.6\columnwidth}
\includegraphics[width=\columnwidth]{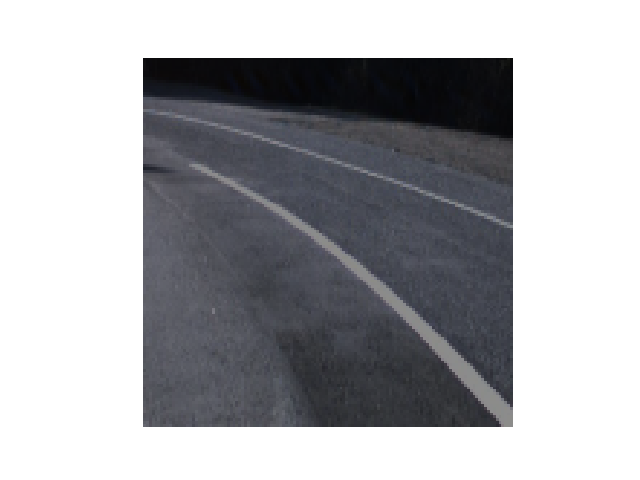}
\caption{Adversarial image, 'right'}
\end{subfigure}

\begin{subfigure}[t]{0.6\columnwidth}
\includegraphics[width=\columnwidth]{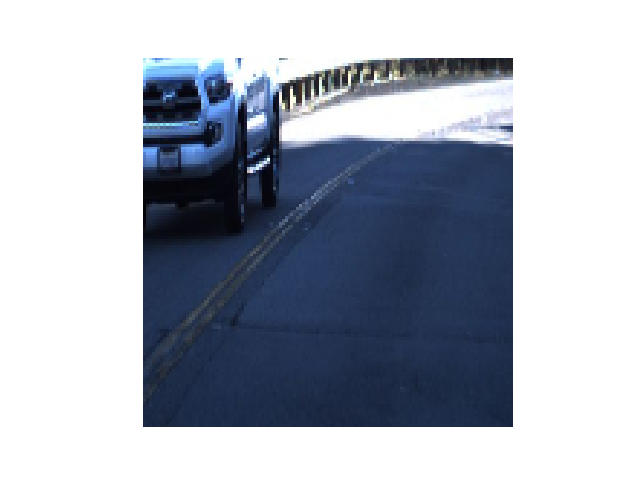}
\caption{Input image,  'right'}
\end{subfigure}
\begin{subfigure}[t]{0.6\columnwidth}
\includegraphics[width=\columnwidth]{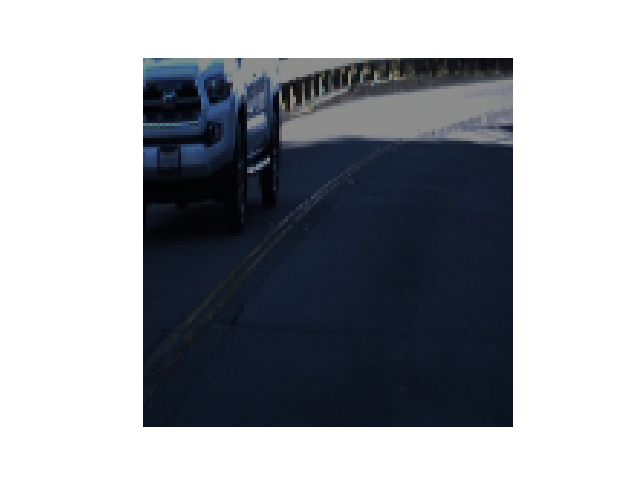}
\caption{Adversarial image, 'straight'}
\end{subfigure}
\begin{subfigure}[t]{0.6\columnwidth}
\includegraphics[width=\columnwidth]{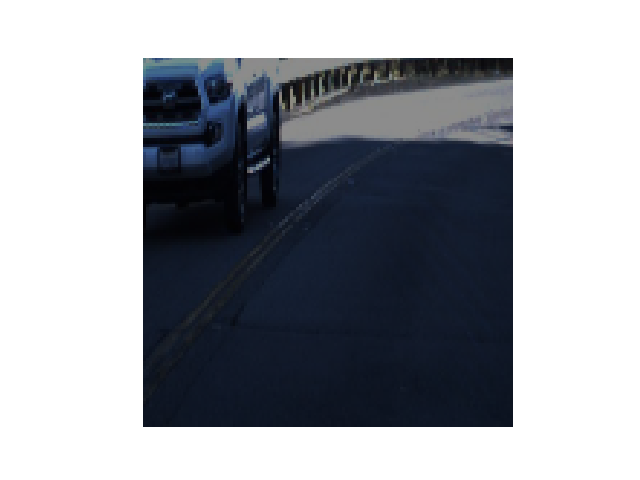}
\caption{Adversarial image, 'left'}
\end{subfigure}
\caption{Adversarial images for the NVIDIA classification model.}
\label{fig:exnvidia}
\end{figure*}

\myparagraph{Data} The training data consists of 33,608 images extracted from the videos provided by the Udacity self-driving car challenge 2. We apply image preprocessing as done by previous work~\cite{UdacitySDC,tian2018deeptest}:  crop the images from the original size ($640 \times 480$ pixels) to $640 \times 280$, and resize the images to $128 \times 128$ pixels. Each datapoint includes the steering angle at the moment the image was captured. The steering angles in the Udacity data set driving log were pre-scaled by a factor of 1/25 (see the histogram of the scaled angles in Figure~\ref{fig:hist} and statistics on the distribution in Table~\ref{tab:angle}).

To assign classification labels, we split the scaled angle values into 3 intervals to obtain  the 3 directions (left, straight, and right). The histogram, as well as discussion with domain experts, motivates the choice of the scaled angle threshold at 0.15, resulting in majority of labels to be straight (70\%), and 15\% of labels to be set as left and right, respectively.

\myparagraph{Training results} We train both models using 10-fold cross validation. The accuracy for classification is high: 90\% for the Epoch model and 86\% for the NVIDIA model.
The hyper-parameters for both models are in Table~\ref{tab:params}. While it is possible to improve the accuracy of the NVIDIA model further by using regularization and parameter tuning, we did not pursue this direction as being an orthogonal goal to our paper's main focus. For the regression problem, we only report results for the Epoch model, on which we obtain MSE of 0.03.




\myparagraph{Attack results for direction prediction} For testing the attack we choose 300 images from all 3 classes, and select the 2 values of the targeted class (different from original class), resulting in 600 adversarial images. We found the optimal value for the attack hyper-parameter $c$ by running binary search for 9 steps with the initial value of $c$ equal to $0.001$. As expected, if there are no constraints on adversary's ability to manipulate images, the adversarial success rate reaches 100\%.  Our goal is to understand the minimum amount of $L_2$ perturbation needed for succeeding at generating adversarial examples.

The attacker success with respect to the amount of perturbation is illustrated in  Figures~\ref{fig:dist1} and~\ref{fig:dist2}. In the Epoch model, a minimum modification to the image (0.82 $L_2$ norm) results in 100\% attack success. However, the amount of perturbation for NVIDIA is higher (121.01 $L_2$ norm). We conjecture the reason to be the additional complexity of the NVIDIA model, resulting in a more robust architecture.

Finally, we study the impact of the attack on the models' performance. The micro-average ROC curves with and without the attack are in Figures~\ref{fig:ROC1} and~\ref{fig:ROC2}, respectively. False positive rate for each class is the number of adversarial images classified as this class. It could be easily seen that the model  performance decreases under attack (for instance, AUC decreases from 1 in the no-attack scenario to 0.62 for 0.75 $L_2$ norm perturbation for the Epoch model).




In Figure~\ref{fig:exepoch} we show examples of original images (left), and two corresponding adversarial images (center and right) for the Epoch model. Similarly we show adversarial images for the NVIDIA model in Figure~\ref{fig:exnvidia}. The images look very similar to the original ones, but they become darker as the majority of pixels incur minor modification (a result of our use of the $L_2$ distance). We thus demonstrate that we can modify images from any source class to any targeted class.
It took on average around 5 and 25 seconds, respectively, to generate adversarial image for the Epoch and NVIDIA models. This result confirms that it takes longer to attack more complex models.

\begin{figure*}[h]
\begin{minipage}{\columnwidth}
\centering
\includegraphics[width=0.7\columnwidth]{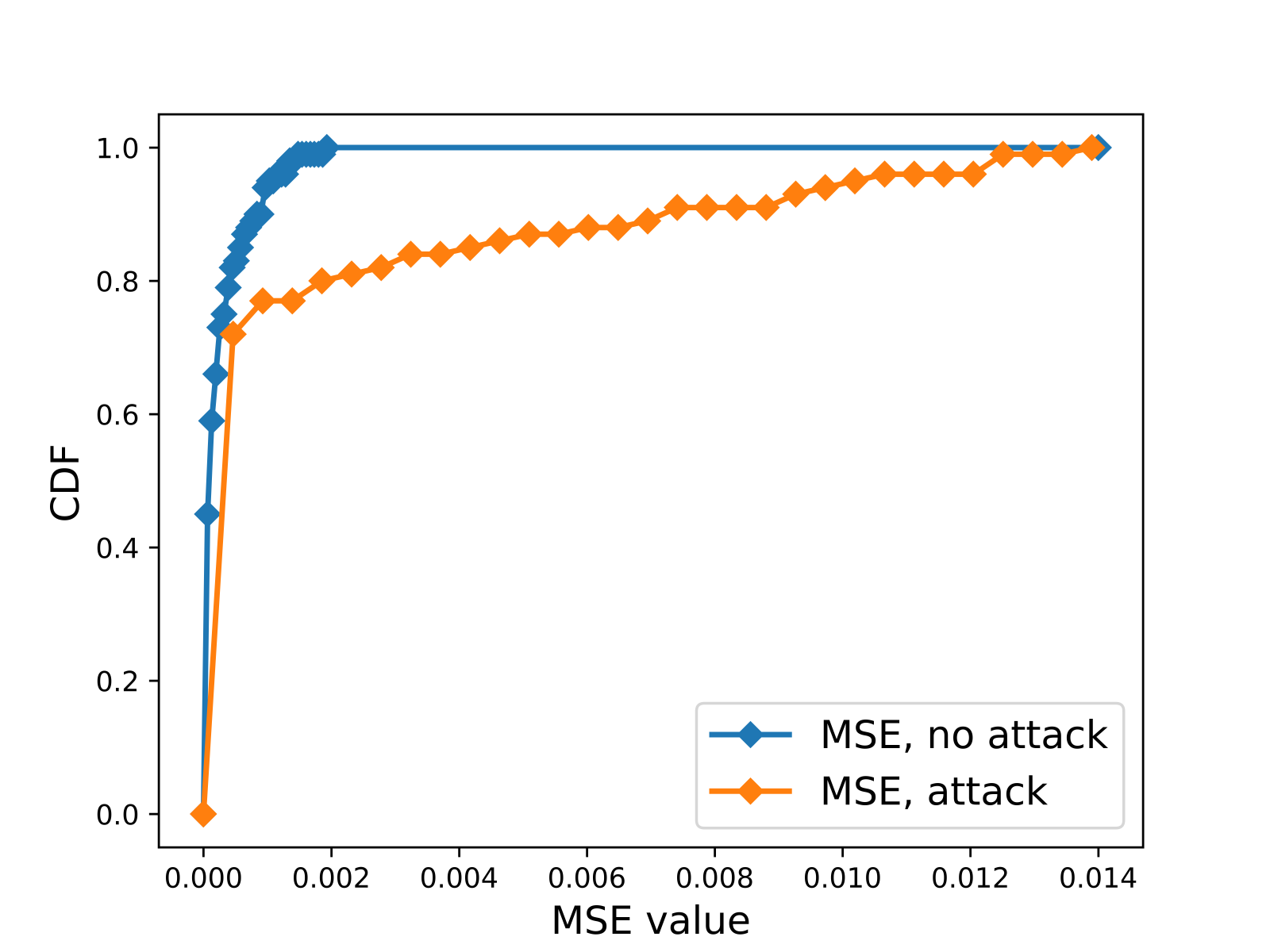}
\caption{MSE CDF}
\label{fig:cdf}
\end{minipage}
\hfill
\begin{minipage}{\columnwidth}
\centering
\captionsetup{type=table}
\begin{tabular}{lll}
\cmidrule(r){1-3}
Percentile   & MSE ratio & Perturbation\\
\midrule
10\%    &  1.19 &  0.007\\
25\% & 1.38 & 0.02 \\
50\%    &  2.43 & 0.05 \\
75\%     &   6.31 & 0.29\\
90\%     &   20.88 & 0.57\\
 \bottomrule
\end{tabular}
\caption{MSE ratio and $L_2$ perturbation statistics}
\label{tab:statistics}
\end{minipage}
\end{figure*}



\myparagraph{Attack results for steering angle prediction}
For testing the attack we choose 100 images. We found the optimal value for the attack hyper-parameter $c$ by binary search. As expected, with higher values of $c$ the attacker obtains adversarial images with high MSE value. We found that the value of hyperparameter $c$ equal to 100 results in the most acceptable tradeoff between MSE and amount of perturbation to the image.



In order to study the success of the attack, we calculate the statistics of $L_2$ norm perturbation values and adversarial to legitimate MSE ratio. These are illustrated in Table~\ref{tab:statistics}. We observe that 90\% of adversarial images have perturbation value less than 0.57 $L_2$ norm, which is very small. Additionally, our attack results in significant changes to the MSE of adversarial images. In particular, 10\% of adversarial images have an MSE value more than 20 times higher than the MSE value of the corresponding legitimate image. The maximum ratio of adversarial to legitimate MSE is 69.

\begin{figure}[t]
\centering
\begin{subfigure}[t]{0.95\columnwidth}
\centering
\includegraphics[width=0.7\columnwidth]{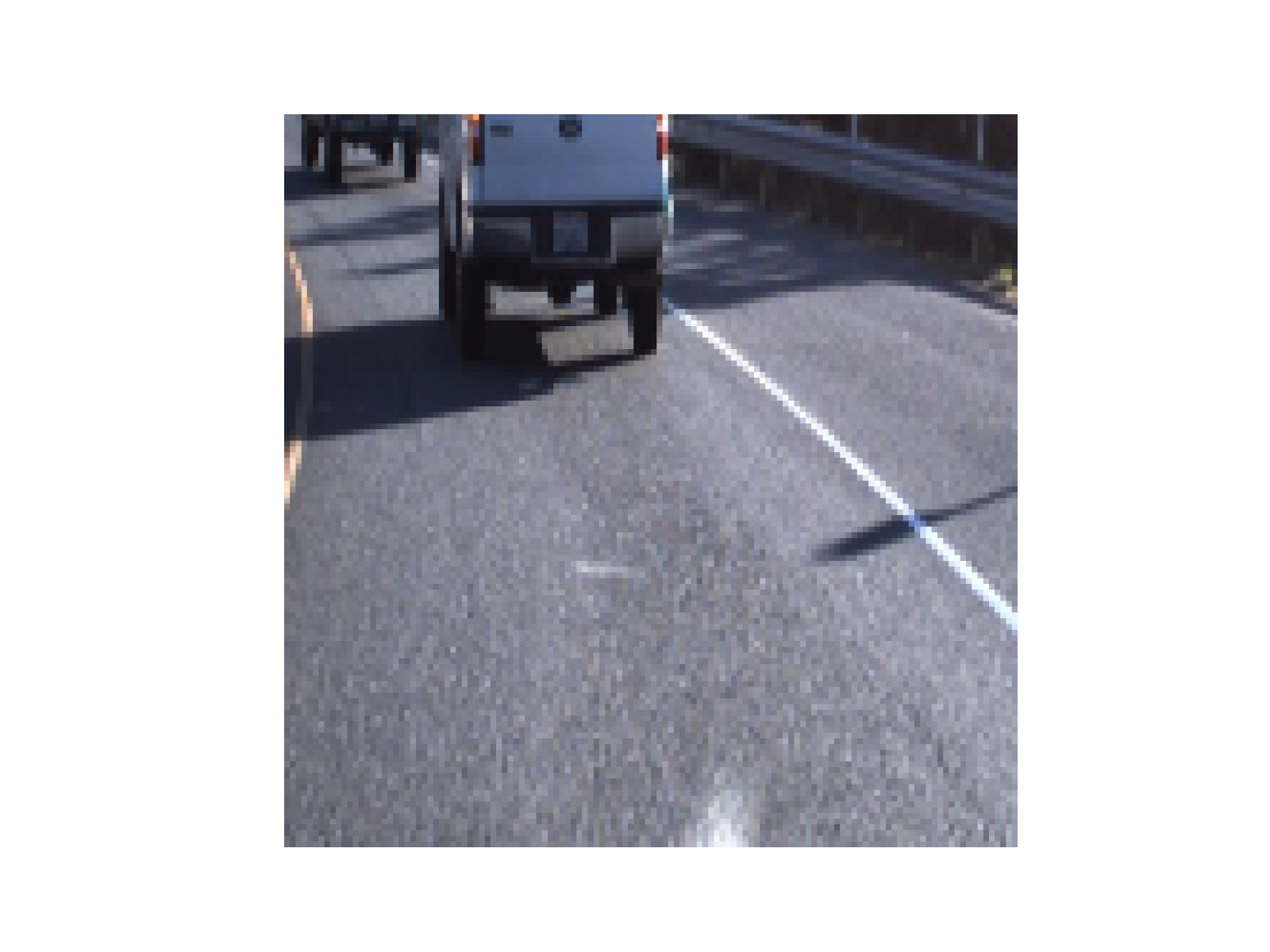}
\caption{Input image, Predicted angle = -4.25, MSE = 0.0016}
\end{subfigure}
\begin{subfigure}[t]{0.95\columnwidth}
\centering
\includegraphics[width=0.7\columnwidth]{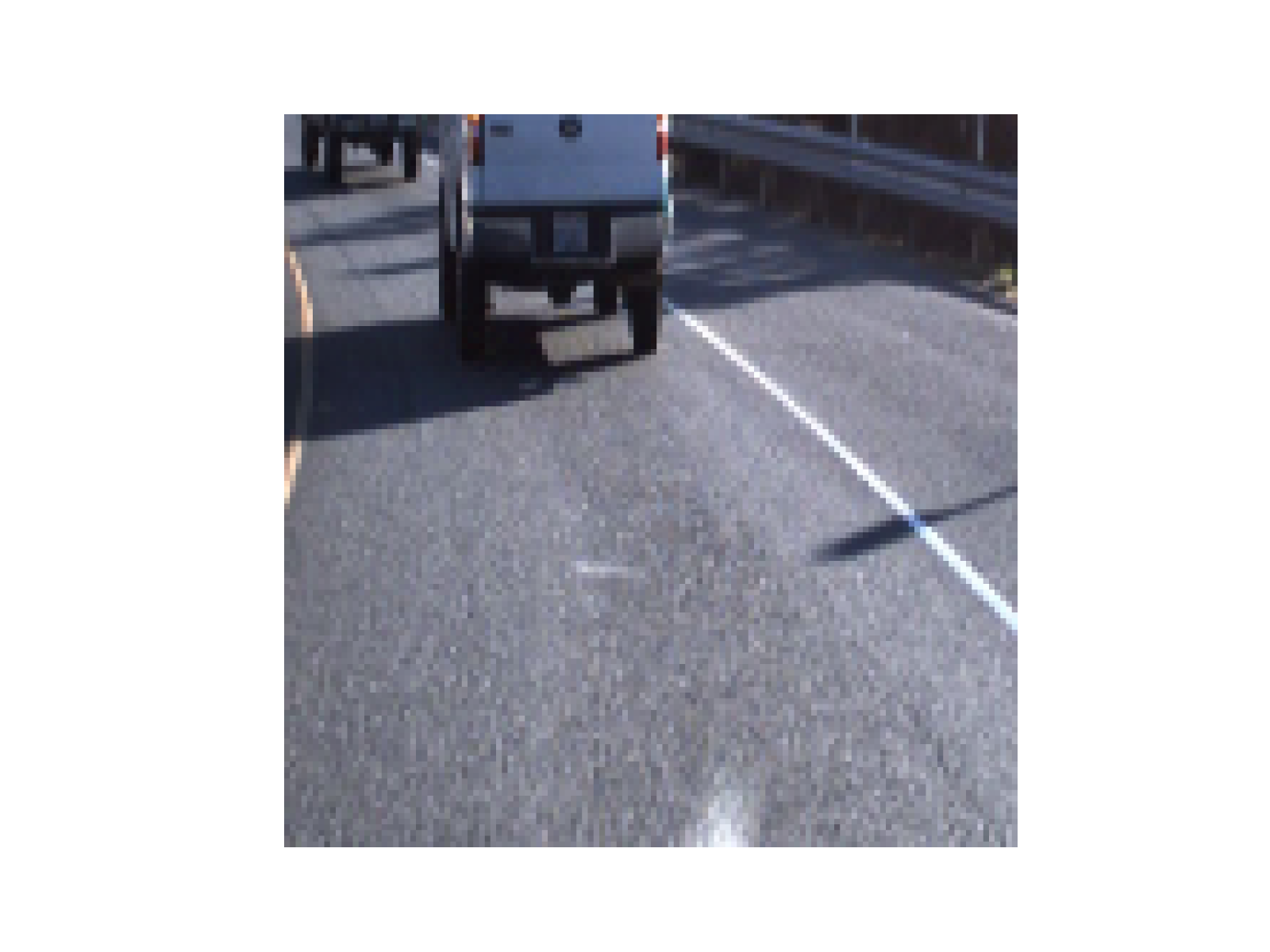}
\caption{Adversarial image, Predicted  angle = -2.25, MSE = 0.05 }
\end{subfigure}
\caption{Adversarial images for the Epoch regression model.}
\label{fig:imgreg}
\end{figure}

Finally, we study the decrease of the model's performance under adversarial attack.
We plot the CDFs of the regression model MSE with and without the attack in Figure~\ref{fig:cdf}. The maximum MSE for the legitimate model is 0.002, while for the adversarial model the maximum MSE reaches 0.014. We show an example of a legitimate image that is transformed into an adversarial image in Figure~\ref{fig:imgreg}. The original steering angle is $-4.25$ degrees, while the adversarial angle results in a value of $-2.25$ degrees, a difference of $47.5\%$.



\section{Conclusion}
The  existence  of  adversarial  examples  limits  the  areas  in
which  deep  learning  can  be  safely applied.
We showed that evasion attacks against neural networks are a real threat for steering angle prediction in autonomous vehicles. With small perturbation to the input images we created adversarial examples that are either mis-classified by the model (in the classification task) or increase the MSE of legitimate images (in the regression task). Defending against these attacks is a challenging open problem.




\section*{Acknowledgement}

This work was supported by a grant from Toyota ITC. 



\bibliographystyle{IEEEtran}
%
%
%


\bibliography{refs}

\end{document}